\documentclass[letterpaper, 10 pt, conference]{ieeeconf}
\IEEEoverridecommandlockouts    
\usepackage[textwidth=4cm,disable]{todonotes}

\usepackage{graphics}           
\usepackage{times}              
\usepackage{amsmath}            
\usepackage{amssymb}            
\usepackage{graphicx}
\usepackage{algorithm}
\usepackage[noend]{algpseudocode}
\usepackage{booktabs}
\usepackage{color}

\definecolor{instructioncolor}{rgb}{.5,.5,.5}

\usepackage[font=small]{caption}

\def\secref#1{Sec.~\ref{#1}}
\def\figref#1{Fig.~\ref{#1}}
\def\tabref#1{Tab.~\ref{#1}}
\def\eqref#1{Eq.~(\ref{#1})}

\makeatletter
\usepackage{xspace}
\DeclareRobustCommand\onedot{\futurelet\@let@token\@onedot}
\def\@onedot{\ifx\@let@token.\else.\null\fi\xspace}
\def\eg{e.g\onedot} 
\def\ie{i.e\onedot} 
 
\def\etc{etc\onedot} 
 
\def\etal{{et al}\onedot}
\def\etalcite#1{\etal~\cite{#1}}

\makeatother

\usepackage{array}
\newcolumntype{L}[1]{>{\raggedright\let\newline\\\arraybackslash\hspace{0pt}}m{#1}}
\newcolumntype{C}[1]{>{\centering\let\newline\\\arraybackslash\hspace{0pt}}m{#1}}
\newcolumntype{R}[1]{>{\raggedleft\let\newline\\\arraybackslash\hspace{0pt}}m{#1}}

\def\argmax{\mathop{\rm argmax}}

\usepackage{subfigure}
\usepackage{siunitx}

\title{\LARGE \bf \method: Robust Localization using Navigation Signs and Public Maps
}

\author{Nicky Zimmerman\dag  \and Joel Loo \dag \and  Ayush Agrawal\ddag \and David Hsu\dag
\thanks{\dag Smart Systems Institute, National University of Singapore
}%
\thanks{\ddag Northeastern University
}
} 

\usepackage{xspace}

\newcommand*{\obs}[1][]{z_t^{#1}}
\newcommand*{\signloc}[1][]{l^{#1}}
\newcommand*{\signdir}[1][]{D^{#1}}

\newcommand{\navcues}{\mathcal{C}}
\newcommand{\map}{m}
\newcommand{\state}{x_t}
\newcommand{\motion}{u_t}
\newcommand{\navgraph}{\mathcal{G}}
\newcommand{\nodes}{\mathcal{V}}
\newcommand{\edges}{\mathcal{E}}
\newcommand{\statenode}{v}
\newcommand{\stateangle}{\theta}
\newcommand{\dirs}{\mathcal{D}}
\newcommand{\method}{\textbf{SignLoc}}

\newcommand{\jem}{Mall}
\newcommand{\ntfgh}{Hospital}
\newcommand{\nus}{Campus}

\DeclareMathOperator{\topk}{\text{argmax}_k}

\def\weblink{\urllink[pre = \bgroup\bf, post = \egroup]}

\begin{document}  

\makeatletter
\let\@oldmaketitle\@maketitle
\renewcommand{\@maketitle}{\@oldmaketitle%
    \centering
    \begingroup
    \setcounter{figure}{0}
    \captionsetup{type=figure}
    \includegraphics[width=1.0\linewidth]{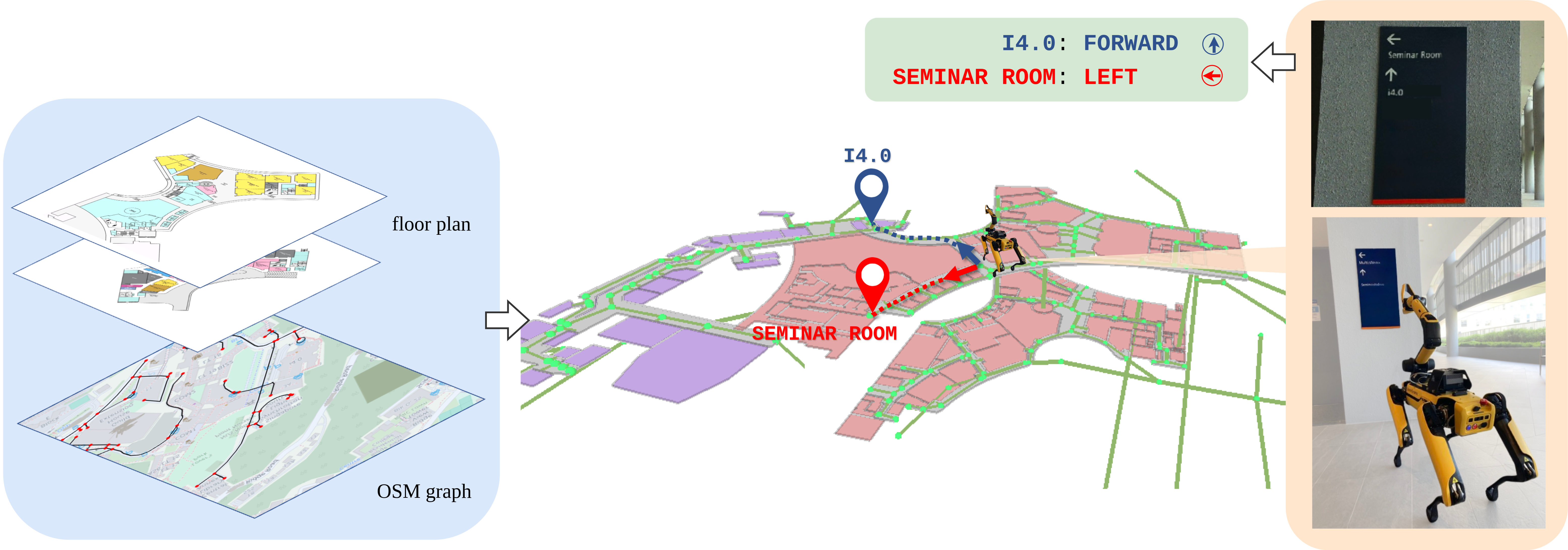}
    \captionof{figure}{Global localization using \method{} with a Boston Dynamics Spot robot. The robot detects navigational signs and extracts directional and locational cues. To localize, it matches these cues to a large-scale, indoor-outdoor navigation graph, constructed from publicly available maps. 
    }
    \vspace{-10pt}
    \label{fig:motivation}
    \endgroup

}
\makeatother

\maketitle

\thispagestyle{empty}
\pagestyle{empty}


\begin{abstract}
Navigation signs and maps, such as floor plans and street maps, are widely available and serve as ubiquitous aids for way-finding in human environments. 
Yet, they are rarely used by robot systems. This paper presents \textit{\method{}}, a global localization method that leverages navigation signs to localize the robot on publicly available maps---specifically floor plans and OpenStreetMap (OSM) graphs--without prior sensor-based mapping. \method{} first extracts a navigation graph from the input map. It then employs a probabilistic observation model to match directional and locational cues from the detected signs to the graph, enabling robust topo-semantic localization within a Monte Carlo framework. 
We evaluated \method{} in diverse large-scale environments: part of a university campus, a shopping mall, and a hospital complex. Experimental results show that \method{} reliably localizes the robot after observing only one to two signs.

\end{abstract}

\section{Introduction} \label{sec:intro}






Localizing and navigating in the open world remains a challenge for robots due to the diversity and complexity of human environments. Prior work enables robust localization, particularly for small-scale indoor environments, by exploiting geometric features~\cite{zhang2014rss} along with some semantic features like objects~\cite{hosseinzadeh2018arxiv} or in-the-wild text~\cite{zimmerman2022iros}. However, these face challenges scaling up to large, heterogeneous environments, particularly those spanning both indoors and outdoors. Existing methods also typically require in-situ sensor data to build representations for localization, necessitating an additional pre-deployment mapping step.

Humans have addressed these challenges by designing wayfinding aids such as \textit{navigational signs} and \textit{publicly available human-centric maps}---\eg, floor plans and street maps. These two aids are complementary, with both encoding symbolic locational and coarse directional information about manmade environments. Signs are informative features capturing 
global information about places and their relative locations. They can be directly matched to features in existing human-centric maps, enabling localization without pre-deployment mapping. Since they are ubiquitous both indoors and outdoors, and are purposefully placed in large-scale spaces to support human navigation, they are ideal semantic features for localizing in large-scale, heterogeneous scenes. However, both signs and human-centric maps remain under-utilized sources of information in robot systems. 

We present \method{} (\figref{fig:motivation}) a global localization approach that leverages navigational signs and publicly available human-centric maps---namely floor plans for indoor spaces and OpenStreetMap (OSM) graphs for outdoor areas. Our approach rests on two key observations. First, human-centric maps encode key locations and the paths between them in a coarse but directionally accurate way, which can be extracted in the form of a \textit{navigational graph}. Second, navigational signs often indicate the \textit{direction of shortest travel} to a given location. This property is a distinctive feature for localization, as it lets us narrow down the set of possible locations for the sign by comparing the sign's directions with the directions of shortest paths computed from all candidate locations in the navigational graph.

Our main contribution is a global localization approach that uses navigational signs as spatial semantic features to localize with respect to a topo-semantic navigational graph extracted from floor plans and OSM graphs. We build on existing work in sign understanding~\cite{agrawal2025arxiv}, and propose a probabilistic observation model to match directional cues extracted from signs, with the navigational graph. This model is used for global localization with a particle filter. We also propose an approach that extracts and stitches together navigational graphs from floor plans and OSM graphs.



In sum, we make three key claims:
Our approach is able to
(i) Extract navigational information from signs;
(ii) Utilize human-centric map priors in the form of floor plans and OSM graphs to establish a unified world representation;
(iii) Localize robustly in large human-centric environments, with seamless transition between outdoor and indoor spaces.
Experiments across diverse large-scale environments---ranging from a mall to a campus to a hospital---support these claims. In particular, we find that \method{} reliably localizes the robot after observing only one to two signs.


\section{Related Work} \label{sec:related}  


Localization for mobile robots has been studied extensively~\cite{thrun2005probrobbook, ullah2024csr}, and geometric approaches gained substantial popularity~\cite{leonard1991tro, fox1999aaai, dellaert1999icra}. In the recent decade, there is a noticeable shift towards utilizing semantic cues~\cite{rosinol2020icra, zimmerman2023ral, cadena2016tro} and considering new map representations~\cite{rosinol2020arxiv, bavle2022ral, wiesmann2023ral}.

\begin{figure*}[t]
  \includegraphics[width=0.95\linewidth]{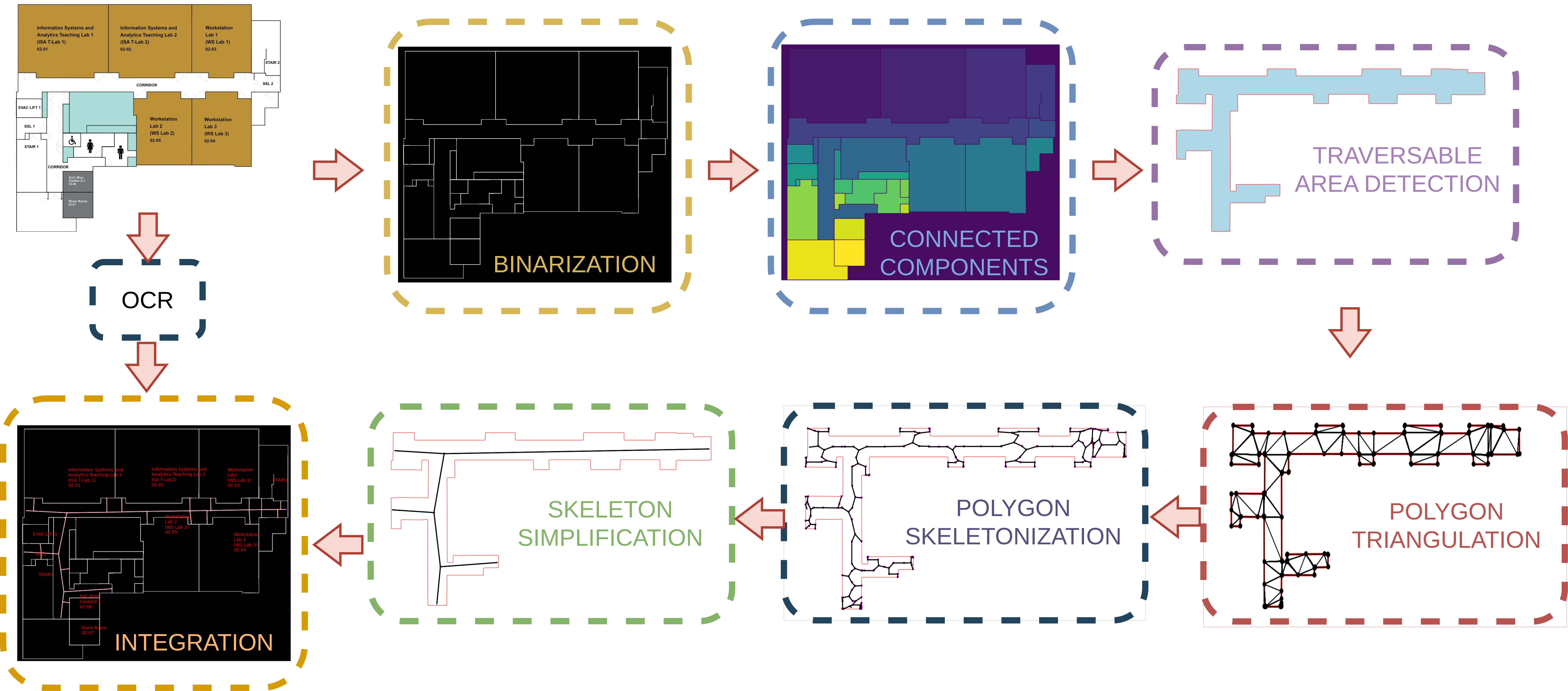}
  \caption{A diagram of the map extraction pipeline, starting with a venue map or a floor plan, and extracting the navigational graph. After the venue map is binarized and the text and symbols are extracted, computational geometry methods are applied to extract the traversable nodes and the place nodes of the navigational graph. Then symbol and text information is integrated into the graph, connecting portal nodes and assigning labels to place nodes. }
  \label{fig:mappingflow} 
\end{figure*}

\textbf{Semantic cues.} 
The need for reliable and informative, yet compact features drives the use of semantic cues in localization. Recent works explored various cues, from specific high-level features like lane markings~\cite{schreiber2013laneloc}, to broad semantic features like affordances~\cite{loo2024sam}, objects~\cite{zimmerman2023ral} and semantic part understanding of scenes~\cite{kobyshev2014matchfeats}. Signs are navigational aids commonplace in human-oriented spaces. While a large body of work exists on detecting and classifying signs, both in indoor~\cite{almeida2019visi, Wang2013nmahib} and on-road settings~\cite{wang2023nca, zhang2022hccis, hoferlin2009ivs, de1997tie}, the application of signs to localization remains limited.  Cui~\etalcite{cui2021iros} and Zimmerman~\etalcite{zimmerman2022iros} detect and associate textual cues on signs to locations in floor plans, while Escalera~\etalcite{de2003ivc} classify traffic signs for on-road localization. Liang~\etalcite{liang2020iros} goes further by parsing text and direction information from signs, using them to guide local mapless navigation.

We use navigational signs as semantic features for global localization in large-scale, heterogeneous scenes. Specifically, we propose an observation model based on the property that signs often reflect the direction of shortest travel to given locations, to localize on human-centric maps. Unlike Liang~\etalcite{liang2020iros}'s rule-based approach, we follow Agrawal~\etalcite{agrawal2025arxiv} and use VLMs for more open-ended sign understanding.

\textbf{Publicly available, human-centric maps.}
Most robot localization approaches leverage dense maps, \eg occupancy or voxel grids~\cite{moravec1985icra}, built from sensor data taken during a pre-deployment mapping run. In contrast, humans localize and navigate effectively in unseen spaces by using abstract, semantic maps like floor plans, road maps \etc, which are made publicly available to aid wayfinding. Recent works explore harnessing human-centric maps for localization and navigation, with many focusing on floor plans for indoor scenes, and OpenStreetMap (OSM) information for outdoor scenes. Several approaches seek to segment and parse floor plans~\cite{liu2017iccv, zeng2019iccv, kim2020jcce, xu2024ijdar}, but often fail to handle floor plans that deviate from the architectural standard, such as the visitor and venue maps or operate under the Manhattan world assumption. Xie~\etalcite{xie2020icra} proposed to generate a navigational graph from a floor plan to localize and guide the visually impaired, but their map extraction fails to generalize to different floor plans and is not suitable for global localization. Cheng~\etalcite{cheng2014tase} propose a localization approach centered around detecting the topology of a junction, but their hand-crafted classifier can only detect a closed set of topologies and they restrict their approach to ``regular buildings'', a criteria that many public facilities architectures fail to meet. To localize in unseen outdoor scenes, Yan~\etalcite{yan2019ecmr} use building outlines as features, while Ruchti~\etalcite{ruchti2015icra} rely on road network topologies. Other visual place recognition approaches, \cite{vysotska2019ral, panphattarasap2018arxiv, zamir2010eccv} exploited Google Street View for image matching. 

We use publicly available human-centric maps---floor plans and OSM graphs---as sources of prior map information to globally localize in unseen environments. We propose a method to extract and stitch a unified navigational graph from floor plans and OSM graphs, which can scale to large areas and span both indoor and outdoor scenes.

\section{\method} \label{sec:approach}

We use navigational signs to globally localize with respect to publicly available human-centric maps---floor plans and OpenStreetMap (OSM) graphs. We decompose this problem into three main steps: (i) \textit{map extraction}, (ii) \textit{sign parsing}, and (iii) sign-centric localization. Map extraction (\secref{sec:navgraph}) extracts and stitches together a unified navigational graph $\navgraph$ from floor plans and OSM graphs. Sign parsing (\secref{sec:sign_parsing}) extracts navigational cues---associated textual and directional features---from navigational signs. Sign-centric localization (\secref{sec:localization}) describes our novel observation model for matching navigational cues to $\navgraph$ for global localization.

\subsection{Navigational Graph Construction based on Priors} \label{sec:navgraph}

We propose an approach to extract and stitch together a single navigational graph $\navgraph$ from floor plans and OSM graphs. Navigational graph construction takes as input floor plans as 2D top-down RGB images, and OSM graphs. The output is the navigational graph $\navgraph = (\nodes, \edges)$, inspired by \cite{yang2015ijgis}. The graph nodes $\nodes$ comprise three types: \textbf{(i)} \textit{intersection nodes} where two or more distinct navigable paths meet (\eg turn in corridor, T-junction \etc); \textbf{(ii)} \textit{place nodes}, which correspond to named regions in the input maps and store the region's textual label as an attribute; \textbf{(iii)} \textit{portal nodes} that connect disjoint areas (\eg doors, lifts, stairs). The graph edges $\edges$ represent connectivity between the nodes. The extracted graph $\navgraph$ need not be metrically accurate, but it should preserve coarse directional information. Specifically, \method{} discretizes directions that edges $e\in\mathcal{E}$ can take into the set of 8 cardinal directions $\dirs$. If metrically accurate OSM maps are input, the $\navgraph$ should be metrically embedded in the coordinate frame of the OSM map. 

We apply computational geometric techniques to obtain graphs from an indoor floor plan. If an OSM graph is present, it can serve as the ``backbone'' of the navigational graph, being used to coarsely embed subgraphs from individual floor plans into a consistent global coordinate frame. This yields a single navigational graph $\navgraph$ spanning from indoors to outdoors.

\textbf{Graph extraction from floor plans.} When floor plans or venue maps are available, the indoor navigational graph can be extracted automatically. While previous works~\cite{xie2020icra} attempted to construct navigational graph from binarized maps in the pixel-space, we work in polygon-space. We skeletonize the binary image and detect enclosed areas using connected components analysis, and we capture these geometries as polygons. The nodes of the extracted, simplified skeleton are intersection nodes, while the enclosed areas are place nodes. We detect traversable spaces, such as corridors and lobbies, by their degree of connectivity to other rooms. For these traversable areas, we use a polygonal skeletonization approach~\cite{kieler2009agile} to extract the centerline on which we should navigate. For the non-traversable spaces, such as enclosed rooms, they are assigned a place node and are associated with a name that is extracted from the map using Paddle-OCR. 

Portal nodes such as doors, lifts and stairs are detected using their name or a symbol in the floor plan. The detections of symbols, as well as removing clutter, text and colors to achieve the binary image that captures the geometry, are performed by querying a VLM. As the results may be imperfect, annotating a navigational graph or editing an existing graph is also available through our GUI application NavGraphApp. The GUI allows users to construct navigational graphs when prior maps are unavailable, too cluttered, outdated or lacking important details.

\textbf{Registration to global coordinate frame.} The procedure detailed above is aimed at a single source map, which usually describes one floor of a building. To address the construction of multi-building navigational graphs given separate floor plans, we propose an alignment process based on OpenStreetMap~\cite{haklay2008pc}. As we store a simplified exterior polygon of each floor in our floor nodes, we can register it against the polygon provided by OpenStreetMap. We find the similarity transform that maximizes the overlapping area between the two polygons. Given polygon $P_{\text{fp}}$ extracted from the floor plan and polygon $P_{\text{osm}}$, we optimize the similarity transform $\boldsymbol{M}$ to maximize the IoU between the two polygons,

\begin{equation}
 \boldsymbol{M}_{\text{opt}} = \argmax_{\boldsymbol{M}} \mathrm{IoU} \big(
 \mathrm{A}(P_{\text{osm}}), \mathrm{A}(\boldsymbol{M}P_{\text{fp}}) \big)
 \label{eq:polyregister}
\end{equation}
where $ \mathrm{A}(\cdot)$ donates a function that computes the area of a polygon and the similarity transform is defined as $\boldsymbol{M}= \boldsymbol{R}\boldsymbol{S} + \boldsymbol{T}$.

For multiple building with multiple floor plans or venue maps, we register each against its OSM counterpart, and transform each navigational graph to a global coordinate system where they are aligned. When the navigational graphs are aligned with the OSM coordinate frame, utilizing the road network to extend the navigational graph becomes trivial. The process of aligning floor plans and venue maps against the OSM polygons essentially embeds the topological graphs in 3D space, enabling also topometric localization.

\begin{figure*}[t]
  \centering
  \includegraphics[width=0.95\linewidth]{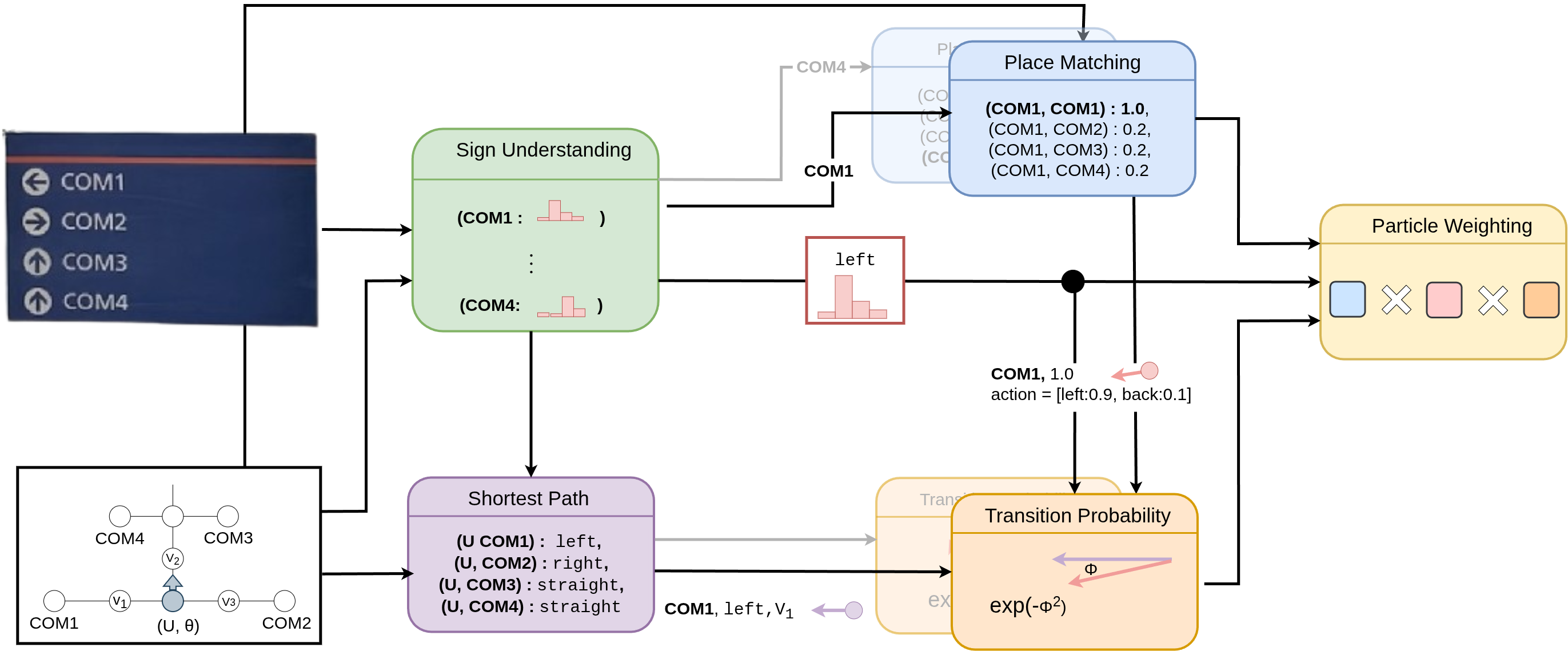}
  
  \caption{An overview of the observation model. The input is the navigational graph and image of a directional sign. The observation model computes how likely it is to observe the sign for each particle with state $(U, \theta)$, and reweighs the particle accordingly.}
  \label{fig:observationmodel} 
\end{figure*}

\subsection{Sign Parsing}
\label{sec:sign_parsing}

The sign parsing module first identifies navigational sign candidates in RGB images, then processes them with a VLM-based sign understanding system~\cite{agrawal2025arxiv} to extract the embedded navigational cues. 

\textbf{Shortlisting candidate signs.} 
We seek to identify candidate signs relevant to localization from incoming RGB images. Specifically, we identify image inputs containing text strings of places that appear in prior maps. This is done using fast text spotting (\eg{} PaddleOCR~\cite{cui2025paddleocr30technicalreport}) to extract observed text, and filter based on the list of place strings extracted from the prior map (\secref{sec:navgraph}). Selected images are passed to an open-set object detector (GroundingDINO~\cite{liu2023grounding}) that is prompted to predict bounding boxes for ``navigational signs'' in the image. This produces candidate signs containing navigational cues useful for localization. 

\textbf{Sign understanding.} 
We adapt the sign understanding pipeline of \etalcite{agrawal2025arxiv} to extract symbolic navigational cues from candidate signs. We extend their definition of navigational cues to tuples $(\signloc, \signdir)$, where $\signloc$ is the location label, and
$\signdir$ is a probability distribution over the set of direction categories, $\dirs$, which discretizes possible orientations into 8 cardinal directions. Each element of $\signdir\in\mathbb{R}^8$ represents the likelihood that $\signloc$ corresponds to that directional category. Sign understanding infers the list of navigational cues present in the sign, $\navcues{} = \{(\signloc[1], \signdir[1]), \dots, (\signloc[J], \signdir[J]) \}$. To handle diverse sign layouts seen in the open world, the pipeline prompts a VLM to generate $\navcues$. In particular, the VLM is prompted iteratively, and its responses are used to empirically construct the uncertainty distribution $\signdir$.

\subsection{Sign-Centric Localization}
\label{sec:localization}

\begin{figure}[t]
  \includegraphics[width=0.95\linewidth]{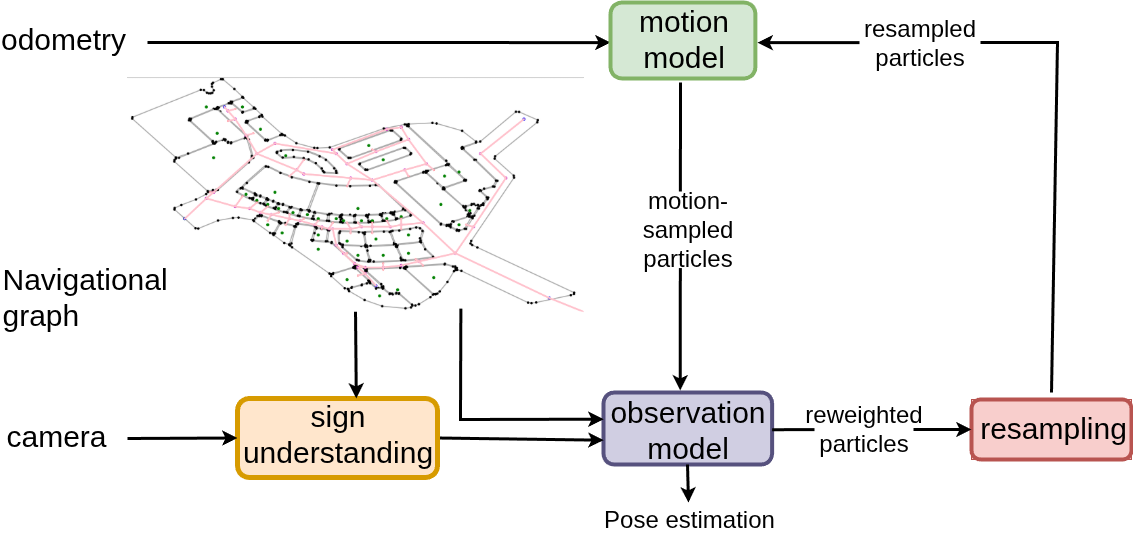}
  \caption{An overview of our localization approach, based on our navigational graph and utilizing navigational cues from signs.}
  \label{fig:localizationflow} 
\end{figure}

The localization module globally localizes the robot with respect to the map obtained in \secref{sec:navgraph}, using navigational cues from signs, $\navcues{}$, obtained in \secref{sec:sign_parsing}. Our approach builds on the Monte Carlo localization framework~\cite{dellaert1999icra}, which estimates the robot's state $\state{}$ in a given map $\map{}$ based on motion priors $\motion{}$ and observations $\obs{}$. In this particle filter formulation, when motion priors are available, the particle's state is updated via the \textbf{motion model}. The \textbf{observation model} reweighs the particle based on the likelihood of the observation given the particle's state. During the \textbf{resampling} step, particles are sampled based on their weights, and particles with higher weight are more likely to be selected, while weak hypotheses perish. An overview of the localization can be seen in \figref{fig:localizationflow}.

In our sign-centric localization, the map $\map{}$ takes the form of the navigational graph $\navgraph{} = (\mathcal{V}, \mathcal{E})$. We define the robot's state $\state = (\statenode{}, \stateangle{})$, where $\statenode{}\in\mathcal{V}$ and $\stateangle{}$ is the robot's heading. The observation $\obs{}$ are the navigational cues $\navcues{}_t$, obtained from signs. Specifically, we design a particle filter for global localization, and describe the associated observation and motion models.

\textbf{Observation model.}
Given observed navigational cues $\obs{} = \{\obs[1], \dots, \obs[J]\} = \{(\signloc[1], \signdir[1])_t, \dots, (\signloc[J], \signdir[J])_t\}$, the observation model (\figref{fig:observationmodel}) is:

\begin{equation}
p(\obs{} \mid \state{}, \map{}) = p(\navcues{}_t \mid \state{}, \navgraph{}) = \prod_{j=1}^J p(\obs[j]{} \mid \state{}, \navgraph{})^{\frac{1}{J}}
\end{equation}

Specifically, the observation model is the geometric mean of observation likelihoods for each $\obs[j] = (\signloc[j]_t, \signdir[j]_t)$. Each likelihood models how likely it is for a location corresponding to $\signloc[j]_t$ to be found in a given direction $d$, with respect to the robot's current location at node $\statenode_t\in\mathcal{V}$ with heading $\stateangle_t\in [-\pi, \pi)$:

\begin{multline}
    p(\obs[j] \mid \state, \navgraph) = p((\signloc[j]_t, \signdir[j]_t) \mid (\statenode_t, \stateangle_t), \navgraph) \\
    = \sum_{u \in \mathcal{V}_{\text{top-}k}} p(\signloc[j]_t = \text{label}[u]) \sum_{d\in\mathcal{D}} p(\text{toward}(u) = d \vert (\statenode_t, \stateangle_t), \navgraph)
\end{multline}

Due to ambiguous place names or parsing errors, the extracted label $\signloc[j]_t$ may not match exactly with any node labels, or may match with multiple node labels in $\navgraph$. To handle this, we consider the top-$k$ most similar node labels, \ie $\mathcal{V}_{\text{top-}k} = \topk\limits_{v\in\mathcal{V}}\limits p(\signloc[j]_t = \text{label}[v])$, where $p(\signloc[j]_t = \text{label}[v])$ is a normalized fuzzy matching score based on Levenshtein distance~\cite{miller2009levenshtein}.

We assume that signs indicate the \textit{direction of shortest travel} to a goal. Thus, for each matched node $u\in\mathcal{V}_{\text{top-}k}$, we want to model the likelihood that moving in direction $d$ takes us along the shortest path from $\statenode_t$ toward $u$. In \method{}, the set of directional categories for edges, $\dirs$, is the 8 cardinal directions corresponding to the relative headings $\{0, \frac{\pi}{4}, \frac{\pi}{2}, \frac{3\pi}{4}, \pi, -\frac{3\pi}{4}, -\frac{\pi}{2}, -\frac{\pi}{4}\}$ when the robot is in a pose facing the sign head-on. This likelihood is modelled as:

\begin{equation}
    p(\text{toward}(u) = d \mid \state, \navgraph)
    = p(d_{\text{edge}} =d) \exp{(-(d_{\text{edge}}\cdot d_{\text{act}})^2)}
\end{equation}

$d_{\text{edge}}\in\dirs$ is the normalized direction vector of the edge $(\statenode_t, w)$ in the world frame, where $w$ is the next node along the shortest path from $\statenode_t$ toward $u$. $d_{\text{act}}$ is a world frame normalized direction vector indicating the direction the robot should face to reach $w$, obtained by composing the direction $d\in\dirs$ with the robot's current heading $\stateangle_t$. The weight $p(d_{\text{edge}} = d)$ is the prior probability that the direction of shortest travel toward $u$ corresponds to direction $d\in\dirs$, and is obtained from the distribution provided by sign parsing, $\signdir[j]_t$. $-d_{\text{edge}}\cdot d_{\text{act}}$ computes the disparity between the direction of shortest travel and the robot's current heading using cosine similarity.


\textbf{Resampling.}
After the particles have been re-weighted, we resample the particles using a mixture of reciprocal sampling~\cite{prorok2012icra} and low-variance sampling~\cite{thrun2005probrobbook}, with an efficient sample size criteria~\cite{arulampalam2002tsp} of $\frac{N}{2}$, where $N$ is the number of particles. We assume sign detections happen in proximity to a node rather than an edge, therefore sampling new particles close to nodes. The nodes and orientations are sampled proportionately to weight of their particles, meaning a highly likely particle $(u, \theta)$ would be sampled multiple times with some Gaussian noise, while particles with low weight would perish.

\textbf{Motion models.}
We provide two motion models, for purely topological localization and for topometric localization. In the topological setting, the motion model is based on a discretized action space, comprising motions in the 8 cardinal directions. We use these discretized actions as the motion prior $\motion$, to update our state $\state$. The robot state is updated only when reaching the end of the current traversed edge. 

We also extend the framework to handle the topometric setting, where the navigational graph $\navgraph$ can be embedded in 3D Euclidean space. In this case, we track the robot pose $p_t = [x, y, z, \theta]$ and use the robot's odometry as $u_t$, sampling from the proposal distribution $p(p_t \mid p_{t-1}, u_t)$. $p_t$ can be discretized to obtain the state $\state = (\statenode_t, \stateangle_t)$.




\begin{figure*}[t]
  \centering
  \includegraphics[width=0.95\linewidth]{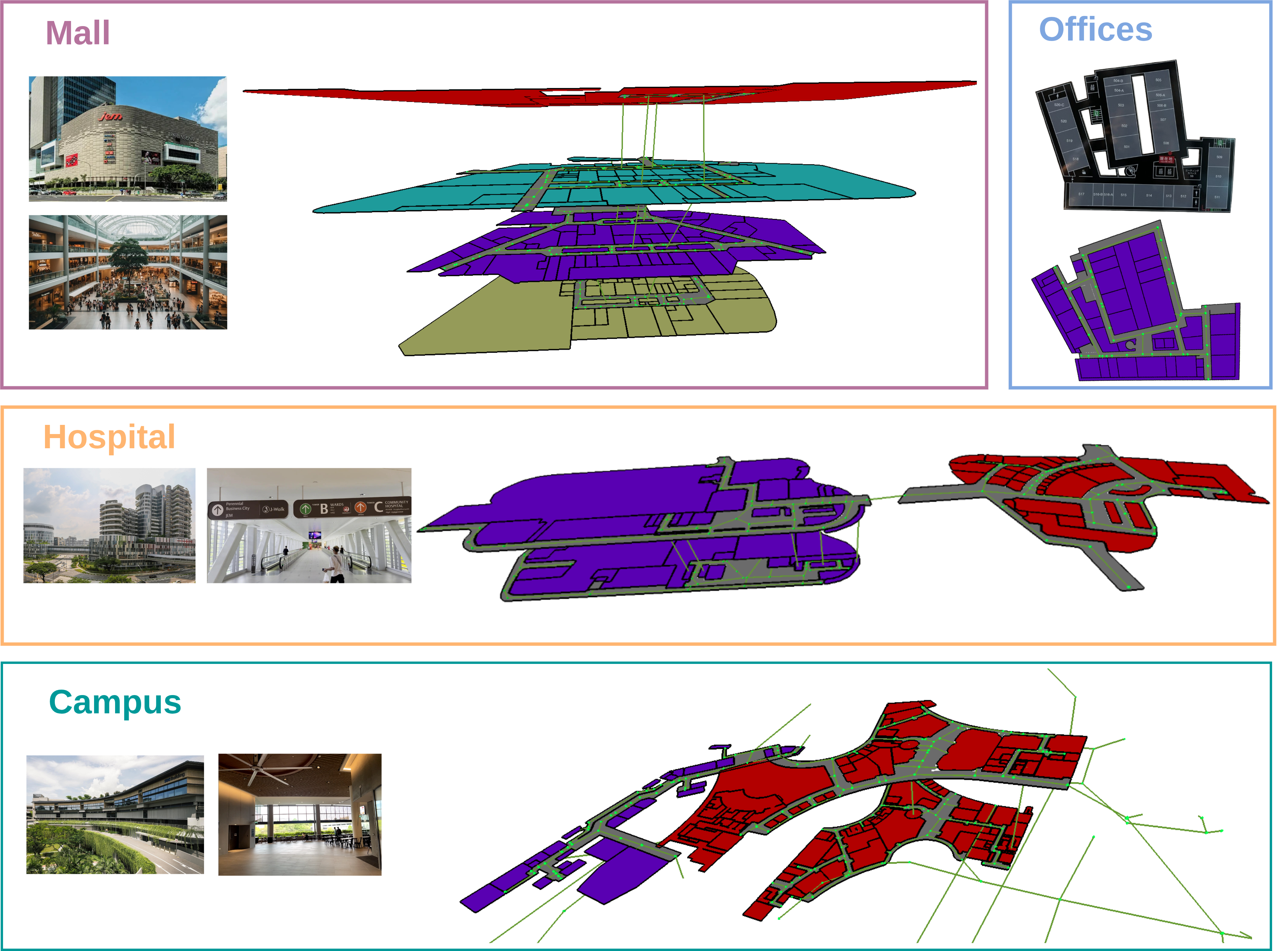}
  
  \caption{Qualitative results of our map extraction approach with multi-floor, multi-building navigational graphs extracted from publicly available maps. 
  In the top right (offices) is our constructed navigational graph for the floor plan used by Xie~\etalcite{xie2020icra}.}
  \label{fig:mappingresults} 
\end{figure*}

\section{Experimental Evaluation} \label{sec:exp}

%
In this work, we propose a novel localization approach that exploits navigational cues from signs, enabling deployment of robots in new environments without prior sensor-based mapping. We conduct experiments to verify the performance of our method. The results of our experiments also support our key claims, which are:
(i) Extract navigational information from signs;
(ii) Utilize various sources of priors to establish a unified world representation;
(iii) Localize robustly in large human-centric environments, with seamless transition between outdoor and indoor spaces.

\subsection{Experimental Setup}
To evaluate the performance of our approach, we recorded a dataset capturing indoor and outdoor public spaces, including a hospital, a mall and a university campus. The tested scenarios included multi-story, multi-building global localization based on publicly available maps, such as OpenStreetMap, floor plans and venues maps. The dataset was collected using a Boston Dynamics Spot and a hand-held setup for places where it was not possible to access with the Spot, and included only odometry and RGB camera stream. 




\subsection{Sign Understanding}\label{sec:signunderstandingeval}

\begin{table}[t]
  \caption{Evaluating sign understanding across 3 different environments.}
  \centering
  \footnotesize
  \begin{tabular}{@{\extracolsep{4pt}}cccc@{}}
    \toprule
    &\textbf{Cue Precision}&\textbf{Cue Recall}&\textbf{Sign Accuracy} \\
    \midrule
    \jem &0.824 &0.847 &0.692 \\
    \ntfgh & 0.830  &0.849 &0.428 \\
    \nus &0.820  &0.837 & 0.583 \\
    \bottomrule
  \end{tabular}
  \label{tab:sign_eval}
\end{table}


We evaluate sign understanding on 10 trajectories across the 3 test environments. The trajectories have human-annotated ground truth navigational cues for each sign in each trajectory. The cue precision/recall metrics are adapted from the sign recognition metrics in \cite{agrawal2025arxiv}, and reflect accuracy of inferring navigational cues, aggregated over all signs in the 10 trajectories. Cue precision is the fraction of accurately inferred cues out of the total number of cues inferred by the sign understanding pipeline, while cue recall is the fraction of accurately inferred cues out of the total number of ground truth annotated cues. We also compute sign accuracy, which denotes the fraction of signs across all 10 trajectories which are perfectly parsed. We define a sign to be perfectly parsed if the set of inferred navigational cues for that sign exactly matches the annotated ground truth.

\tabref{tab:sign_eval} shows that the sign understanding pipeline performs consistently well on navigational signs across diverse environments. The main failure mode observed is incorrect direction category prediction due to complex sign structure in the wild, which may make association of a textual label to a directional arrow ambiguous. Overall, the high cue precision/recall values indicate strong performance of the pipeline, and suggest that each observed sign has a high proportion of accurately parsed navigational cues. While some inaccuracies in parsing individual cues lead to lower sign accuracy values, \method{} is able to utilize the remaining accurately parsed cues from each sign for localization, enabling strong performance in localization.



\subsection{Map extraction}

We qualitatively evaluate our map extraction on several publicly available maps, including a hospital, a shopping mall and a university campus, as well as office buildings from different countries. Overall, the map extraction was tested on 13 separate floor plans and venue maps, from 7 different buildings.  
We also test the augmentation of our navigational graph with information from OpenStreetMap's road network, and show that the navigational graph can also be extended seamlessly to describe outdoor urban environments. We benchmark against the work of Xie~\etalcite{xie2020icra}, who also attempted to extract a navigational graph from a floor plan. While our approach successfully extracted a navigational graph from the map used by Xie~\etalcite{xie2020icra}, as seen in \figref{fig:mappingresults}, their approach failed to produce navigational graphs from the publicly available maps we used for evaluation and localizaion. 
Our tested locations include multi-floor buildings, the alignment of connected buildings with separate floor plans and the extension of the navigational graph to outdoor spaces, all which were not addressed by Xie~\etalcite{xie2020icra}. The qualitative results of our map extraction approach can be seen in \figref{fig:mappingresults} and \figref{fig:motivation}. We show that the extracted navigational graphs can support robust sign-centric localization in \secref{sec:loceval}.
We observe that the most challenging part for our approach is extracting symbols and text from the maps, despite the recent advances of VLMs and OCR approaches.

\subsection{Localization}\label{sec:loceval}
To estimate the accuracy of the topological localization, we evaluate the localization at the locations when navigational signs are detected. We evaluate our localization approach on public spaces; a hospital, a shopping mall, and a university campus, including multi-floor and multi-building scenarios as well as trajectories traversing both indoor and outdoor areas. We globally localize by uniformly initializing the particles across all the traversable nodes of the navigational graph. Therefore, the number of particles used is different for each navigational graph, with the largest graph initialized with 4448 particles. 

Nine sequences were collected to evaluate the performance of the topological localization, each including 5 sign sighting. Sequence S10 includes 7 signs and is used to evaluate the topometric localization.
Sequences S1-S3 are collected in a shopping mall, and the navigational graph includes 4 retail floors. The trajectories contain the usage of escalators to change floors. Sequences S4-S6 were collected in a hospital, and capture transitions between two buildings, Tower A and Tower B. Sequences S7-S9 were recorded in a university campus, across 2 building and also include floor transitions. Sequence S10 starts outdoors at the basement level and traverses up to level 1, across two buildings, measuring at around \SI{300}{\metre} long. More details about the metadata for the navigational graphs used for localization is included in \tabref{tab:mapmetadata}. For the topological localization, we use actions $\{\uparrow, \leftarrow, \downarrow, \rightarrow, ...\}$ taken at each node for the motion model, while for the topometric localization we rely only on odometry from the robot. 

To evaluate the performance of the approach, we compare the localization estimation at signs sighting locations, where ground truth localization is manually annotated based on the known location of the signs. To determine whether our pose estimation is correct, the predicted node must match the GT node, and the orientation difference must be lower than $\frac{\pi}{4}$. Localization is considered successful if the prediction matched the ground truth before the end of the sequence, and if the localization remained converged until the end of the sequence. As can be seen in ~\tabref{tab:loceval}, the global localization converges to the correct topological node and the correct (discretized) orientation in all sequences. The convergence occurs after the first or second sign sighting in 80\% of the cases, but more challenging sequences can require more signs. For all sequences, once the pose estimation converged, it remained correct until the end of the sequence, resulting in 100\% success rate. 
Given the imperfect sign understanding reported in \secref{sec:signunderstandingeval}, we evaluate the performance of our approach with ground truth annotations for the sign understanding. The localization performance is identical to the performance with the predicted labels, reported in \tabref{tab:loceval}. This shows that our localization approach is robust to perception noise and that the convergence is limited by the information contained in each sign, and the occasional mismatch between the navigational signs and the available maps. 
As our approach can utilize both venue maps and OSM information, it was difficult to compare it with an appropriate baseline. While VPR approaches such as Lalaloc++~\cite{howard2022eccv} can localize in floor plans, they cannot localize outdoors and also are challenged by environments with high geometric symmetry. Approaches such as OrienterNet~\cite{sarlin2023cvpr} can globally localize outdoors with a single RGB image and OSM, but cannot be extended to indoor spaces.

\begin{table}[t]
  \caption{Map metaadata for the navigational graphs that were used for localization.}
  \centering
  \footnotesize
  \resizebox{\columnwidth}{!}{%
  \begin{tabular}{ccccccc}
    \toprule
 & nodes & edges & floors & buildings & ind. area & out. area\\
      \midrule
 {\textbf{\jem}} & 641 & 2416 & 4 & 1 & \SI{51338}{\metre\squared}  & - \\
 {\textbf{\ntfgh}} & 309 & 1200 & 2 & 2 &  \SI{30012}{\metre\squared} &  - \\
 {\textbf{\nus}} & 605 & 2145 & 2 & 2 &  \SI{16905}{\metre\squared} & \SI{160000}{\metre\squared}\\
    \bottomrule
  \end{tabular}
  }
  \label{tab:mapmetadata}
\end{table}


\begin{table}[t]
  \caption{Localization performance for 3 different environments. The metric {i/n} reports at which sign sighting i\textsuperscript{th} of the n visible signs the approach pose estimation converged to the ground truth.}
  \centering
  \footnotesize
  \begin{tabular}{@{\extracolsep{4pt}}cccccccccc@{}}
    \toprule
   \multicolumn{3}{c}{\textbf{\jem}} & \multicolumn{3}{c}{\textbf{\ntfgh}} & \multicolumn{4}{c}{\textbf{\nus}} \\
    \cmidrule{1-3}\cmidrule{4-6}\cmidrule{7-10}
    S1 & S2 & S3 &  S4 & S5 & S6 &  S7 & S8 & S9 & S10 \\
      \midrule
    1/5 & 1/5 & 4/5 & 1/5 & 2/5 & 4/5 & 1/5 & 2/5 & 2/5 & 1/7 \\
  
    \bottomrule
  \end{tabular}
  \label{tab:loceval}
\end{table}


\subsection{Runtime}

To evaluate the approach we used a Jetson Orin, which is mounted onboard of the Spot robot. 
The runtime was measured on the university campus navigational graph, which contains 605 nodes, with 4448 particles. The observation and motion model execute in real-time, as well as the sign detection, with the observation model executing for \SI{0.025}{\second} and the motion model in \SI{0.012}{\second}. When a sign is detected, the parsing requires a VLM query, which takes around \SI{2.5}{\second}.  For increased performance in the sign parsing task, the robot is required to be static to avoid motion blur, and therefore the slight delay due to the VLM query is not very noticeable. Overall, our localization approach can run online on the Spot, executing on the Jetson Orin. A video of the localization on the Spot is provided with the submission, and the open-source implementation of our approach will be published on our GitHub repository\footnote{https://github.com/AdaCompNUS/SignLoc}.





\section{Conclusion} \label{sec:conclusion}
We present \method, a global localization approach that matches directional cues from navigational signs to a navigational graph derived from publicly available human-centric maps---floor plans and OpenStreetMap (OSM) graphs. Because signs are ubiquitous in human environments and encode information closely aligned with these maps, they serve as ideal semantic features for large-scale localization spanning indoor and outdoor spaces. Experiments validate this, showing that \method{} rapidly localizes in diverse routes after observing just one or two signs, and demonstrating its ability to globally localize and track the robot's location on a long trajectory crossing both indoor and outdoor areas.




\bibliographystyle{plain_abbrv}

\bibliography{glorified,new}

\end{document}